\documentclass[10pt,twocolumn,letterpaper]{article}

\usepackage{cvpr}              %

\usepackage[pagebackref,breaklinks,colorlinks]{hyperref}

\usepackage{url}            %
\usepackage{booktabs}       %
\usepackage{amsfonts}       %
\usepackage{nicefrac}       %
\usepackage{microtype}      %
\usepackage{graphicx}
\usepackage{amsmath}
\usepackage{amssymb}
\usepackage{graphics}
\usepackage{pifont}
\usepackage{color}
\usepackage{booktabs}
\usepackage{multirow}
\usepackage{adjustbox}
\usepackage{subcaption}
\usepackage{gensymb}
\usepackage{bm}
\usepackage{algorithm} 
\usepackage{algpseudocode} 
\usepackage[table,dvipsnames]{xcolor}
\usepackage{xspace}
\usepackage{wrapfig}

\usepackage[capitalize]{cleveref}
\crefname{section}{Sec.}{Secs.}
\Crefname{section}{Section}{Sections}
\Crefname{table}{Table}{Tables}
\crefname{table}{Tab.}{Tabs.}

\def\cvprPaperID{817} %
\def\confName{CVPR}
\def\confYear{2022}

\newcommand{\citet}[1]{\cite{#1}}
\newcommand{\citep}[1]{\cite{#1}}

\newcommand{\shenlong}[1]{\textcolor{black}{#1}}

\newcommand{\wc}[1]{\textcolor{black}{#1}}

\newcommand{\joycey}[1]{\textcolor{black}{#1}}

\newcommand{\cbd}[1]{\textcolor{black}{#1}} %

\newcommand{\arxiv}[1]{\vspace{0mm}}

\newcommand{\bX}{\mathbf{X}}

\newcommand{\bo}{\mathbf{o}}

\newcommand{\bp}{{\bm{p}}}
\newcommand{\br}{{\bm{r}}}

\newcommand{\cI}{\mathcal{I}}

\newcommand{\bK}{\mathbf{K}}
\newcommand{\bR}{\mathbf{R}}

\newcommand{\bt}{\mathbf{t}}

\newcommand{\sfm}{S\emph{f}M\xspace}

\makeatletter
\g@addto@macro\@maketitle{
\vspace{-3em}
\begin{figure}[H]
   \setlength{\linewidth}{\textwidth}
\setlength{\hsize}{\textwidth}
\centering
\includegraphics[trim={0cm, 0cm, 0cm, 0.0cm},clip,width=\linewidth]{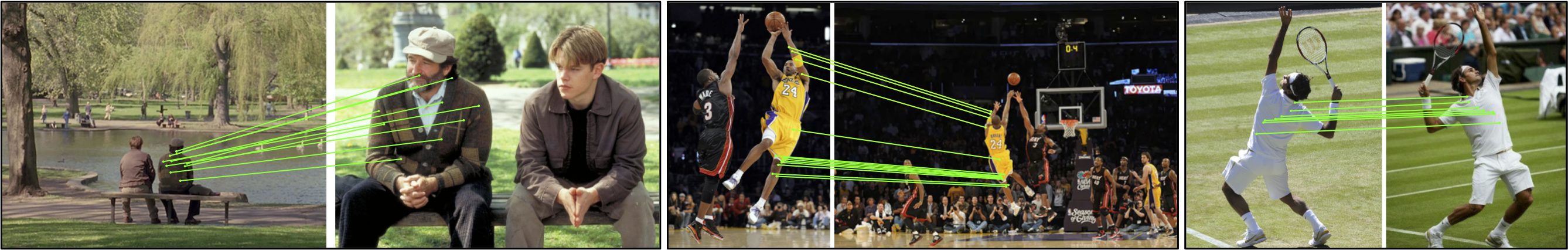}
\vspace{-6.5mm}
\caption{ 
{\bf Can you tell the relationships between these matched pixels?} The head pixel and the face pixel in the leftmost images have completely different semantics and appearances, yet we can still associate them for 3D reasoning. 
Why, and how? 
In this paper, we present a novel concept to establish geometric relationships between pixels even if they are not semantically or visually similar. See Fig. \ref{fig:vc-cars} for cars.
}
\label{fig:teaser-new}
\end{figure}
}
\makeatother

\begin{document}

\title{Virtual Correspondence: Humans as a Cue for Extreme-View Geometry}

\author{%
  Wei-Chiu Ma\textsuperscript{1,4} \quad Anqi Joyce Yang\textsuperscript{2,4,5}  \quad Shenlong Wang\textsuperscript{3}  \quad Raquel Urtasun\textsuperscript{2,4,5}  \quad Antonio Torralba\textsuperscript{1} \vspace*{3pt}\\
  \textsuperscript{1}Massachusetts Institute of Technology \quad \textsuperscript{2}University of Toronto\\ 
  \textsuperscript{3}University of Illinois Urbana-Champaign \quad \textsuperscript{4} Waabi \quad \textsuperscript{5} Vector Institute
}

\maketitle

\begin{abstract}
Recovering the spatial layout of the cameras and the geometry of the scene from extreme-view images is a longstanding challenge in computer vision. 
Prevailing 3D reconstruction algorithms often adopt the image matching paradigm and presume that a portion of the scene is co-visible across images, yielding poor performance when there is little overlap among inputs.  
In contrast, humans can associate visible parts in one image to the corresponding invisible components in another image via prior knowledge of the shapes.   Inspired by this fact, we present a novel concept called virtual correspondences (VCs). VCs are a pair of pixels from two images whose camera rays intersect in 3D.  Similar to classic correspondences, VCs conform with epipolar geometry; unlike classic correspondences, VCs do not need to be co-visible across views. Therefore VCs can be established and exploited even if images do not overlap. 
We introduce a method to find virtual correspondences based on humans in the scene. We showcase how VCs can be seamlessly integrated with classic bundle adjustment to recover camera poses across extreme views.  Experiments show that our method significantly outperforms state-of-the-art camera pose estimation methods in challenging scenarios and is comparable in the traditional densely captured setup.  Our approach also unleashes the potential of multiple downstream tasks such as scene reconstruction from multi-view stereo and novel view synthesis in extreme-view scenarios\footnote{Project page: \url{https://people.csail.mit.edu/weichium/virtual-correspondence/}}.

\end{abstract}
\vspace{-5mm}
\section{Introduction}
\vspace{-2mm}
\label{sec:intro}

Epipolar geometry and correspondence estimation are two keystones of mainstream 3D reconstruction systems. When given a set of RGB images as input, a classic 3D pipeline \cite{longuet1981computer,hartleymultiple} first identifies \emph{co-visible} 3D points across images via pixel-wise visual features and then recovers the spatial relationships among cameras. 
Such a ``golden standard'' framework has experienced huge success in practice and has given birth to numerous applications in robotics, \joycey{AR, VR}, etc. %
The reliance on correspondences, however, makes one ponder: what if the input images have little or no overlap? Does this still work when there are barely any co-visible 3D points in the scene (see Fig. \ref{fig:teaser-new})?

At first thought the answer is no. Predominant correspondence estimators focus on finding pixel pairs that describe the same, co-visible 3D points %
in the scene by matching their visual features. If the viewpoint differences across images are extreme, the pixels will be inherently different and cannot be matched, causing current 3D systems to fail catastrophically.
In contrast, humans can identify where two photographs were taken with respect to the same scene, despite large viewpoint variations.
Such a remarkable capability comes from our prior knowledge of the underlying geometry, which helps us match pixels between images even if their exact correspondences are occluded or invisible in the other image. 
For instance, we know what the front and back of a human body should look like. Therefore, if we see a human face in one image and {the back of a head} %
in the other, we can easily associate them and infer that the two cameras are roughly 180 degrees apart. 
The aim of this paper is to  equip 3D systems with similar abilities.%

Towards this goal, we first ask the following question: %
\emph{do we have to rely on pixels describing the same 3D points to recover camera poses\footnote{We will ignore other primites such as lines or planes for now.}}?
While such (implicit) premises seem to lay the foundation for existing 3D reconstruction algorithms, as we will show in Sec. \ref{sec:method}, the answer is negative. 
Our key observation is that epipolar geometry holds for \emph{arbitrary pixels whose camera rays intersect in 3D}. 
Therefore, as long as we can identify those pixels, 
we can leverage them to recover relative camera poses, {\emph{regardless of whether the pixels are semantically or visually similar or not}.}
{This} interpretation is particularly exciting, as it allows us to go beyond the image space and establish geometric relationships {among pixels} even from {extreme viewpoints}. 

Unfortunately, determining whether two camera rays intersect in 3D often requires camera poses to be known a priori, making the whole process a chicken-and-egg problem.
Our key idea is to exploit prior knowledge of the foreground objects within the scene to break the loop. 
Specifically, we make use of \emph{humans}, arguably one of the most common, salient ``objects'' in images. 
Consider  the images in Fig. \ref{fig:teaser-new}.
If the system has prior knowledge about human shape and pose, 
it will know that a ray shooting through the human back in the leftmost image will intersect with the chest region on its way out. Furthermore, the intersecting chest pixel can be observed in the other image. Thus, we can find a pair of pixels that correspond to two intersecting camera rays with ease. 
\joycey{Note that different from classic correspondences, these two pixels do not depict the same 3D point and thus cannot be found via visual similarities. Since we establish the geometric connection \wc{virtually} by hallucinating a 3D shape, we call them \emph{virtual correspondences} (VCs).}

With this inspiration in mind, we first define virtual correspondences and present a methodology to derive them \joycey{from images containing humans}.
We then showcase how VCs can be seamlessly integrated with the classic bundle adjustment algorithm, resulting in a generalized structure from motion (\sfm) framework that could be applied to \wc{both traditional setup and extreme-view scenarios.}
We evaluate the effectiveness of our approach on the CMU Panoptic dataset \cite{Joo_2015_ICCV,Joo_2017_TPAMI}, 
the Mannequin Challenge dataset \cite{li2019learning}, and multiple challenging in-the-wild images. Our method significantly outperforms prior art in challenging \joycey{extreme-view} scenarios {and is comparable in the conventional, densely overlapping setup}. \wc{Importantly, our estimated poses from extreme viewpoints unleash the potential of multiple downstream applications such as scene reconstruction from multi-view stereo and novel view synthesis {in challenging scenarios}}.

In summary, we make the following contributions:
\begin{enumerate}
    \vspace{-1.5mm}    
    \item We present virtual correspondences, a novel concept for 3D reconstruction algorithms, and establish its geometric connection to existing correspondences.
    \vspace{-2.5mm}
    \item We develop a method to estimate VCs from \joycey{images with} humans and showcase how to integrate them into existing 3D frameworks. The new framework can be applied to a wide range of scenarios \joycey{while} also reduces to the classical \sfm when no VCs are found.
    \vspace{-2.5mm}
    \item We exploit the estimated camera poses for multiple downstream applications and empirically show that 
    our method can extend these tasks to extreme-view scenarios which were previously infeasible.

\end{enumerate}

\section{Related Work}
\vspace{-2mm}
\label{sec:related}
\paragraph{Correspondences:}
Correspondence estimation aims to identify {pixels} that are {projections of the same 3D  point} across multiple images \cite{hartleymultiple,ma2012invitation}. The task has been the cornerstone of various computer vision problems for decades, 
since the pixel-level association %
allows one to recover the structure and motion of the world effectively \cite{longuet1981computer,horn1981determining,beardsley19963d,scharstein2002taxonomy}. 
Prevalent approaches  focus on hand-crafted \cite{sift,brief,brisk,daisy,surf,orb} or learned \cite{wang2016global,autoscaler,ono2018lf,superpoint,wang2019learning,dusmanu2019d2,wang2020caps} robust \emph{visual} features that can distinguish one pixel from the others in diverse scenarios. 
While impressive performance has been achieved  %
 \cite{superglue,sun2021loftr}, 
these methods fall short when there is little overlap among input images, as  there are hardly any co-visible 3D points.
Semantic correspondence estimation \cite{proposalflow,flowweb,zhou2016learning,choy2016universal,kim2017fcss,scnet,scops}, on the other hand, focuses on detecting pixels with specific semantics (\eg, human facial keypoints). With the help of domain knowledge, they are usually more robust to variations in viewpoint, appearance, \wc{and sometimes even occlusions} \cite{hu2016bottom,cao2019openpose}.
Unfortunately, they still require a set of semantic keypoints to be co-visible across multi-view images to enable 3D reconstruction. 
In contrast, our novel virtual correspondences do not have these constraints. 
VCs can be the projection of different 3D points and can have completely different appearances and semantics (\eg, chest pixel \emph{v.s.} back pixel). This allows us to establish geometric relationships among pixels even when the input images have no co-visible 3D points.

\begin{figure}[t]
\vspace{-5mm}
\centering
\includegraphics[trim={0, 0.3cm, 0, 0cm},clip,width=0.98\linewidth]{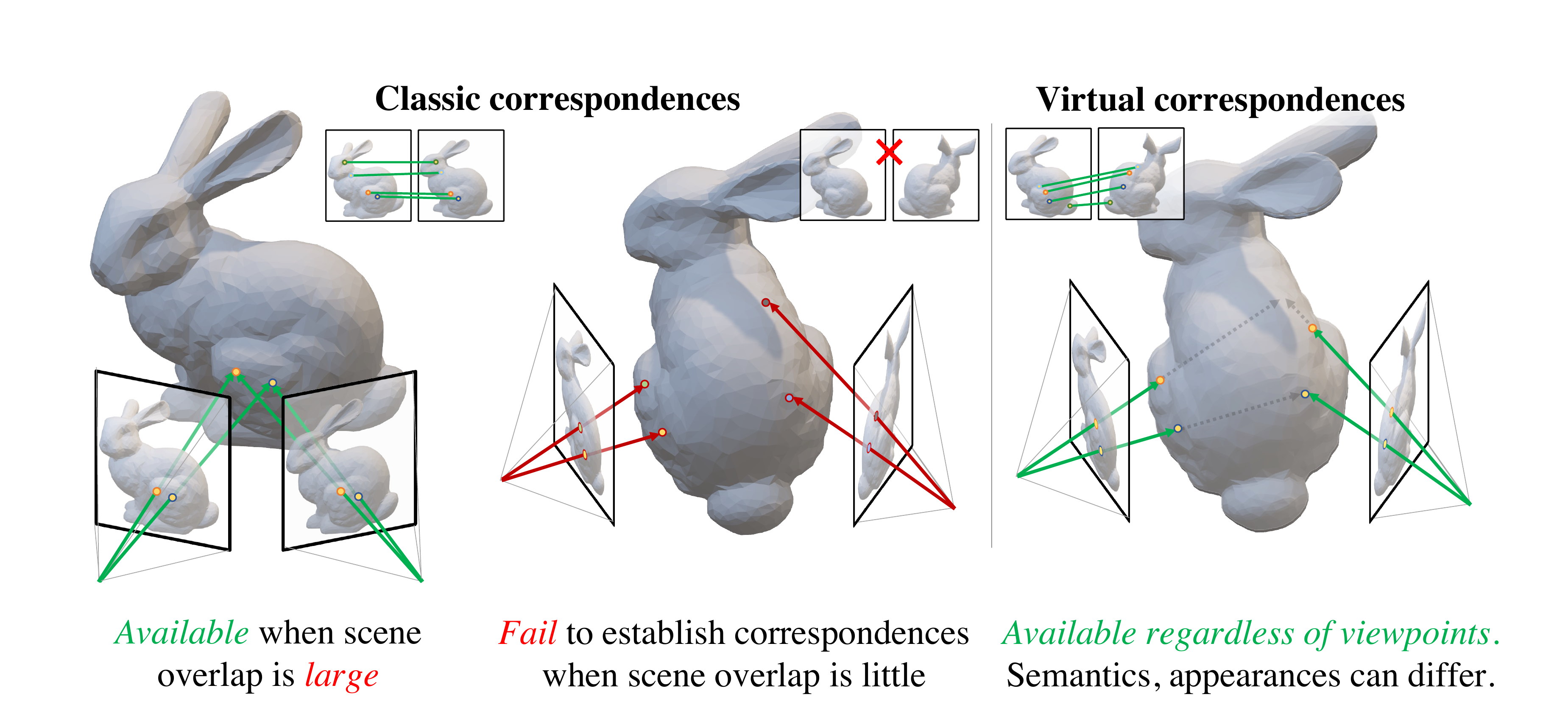}
\vspace{-3mm}
\caption{\footnotesize \textbf{Classic correspondences vs. virtual correspondences.} 
}
\label{fig:vc-bunny}
\vspace{-5mm}
\end{figure}

\begin{figure*}
\vspace{-2mm}
\centering
\includegraphics[trim={0, 0.2cm, 0.5cm, 0cm},clip,width=0.95\linewidth]{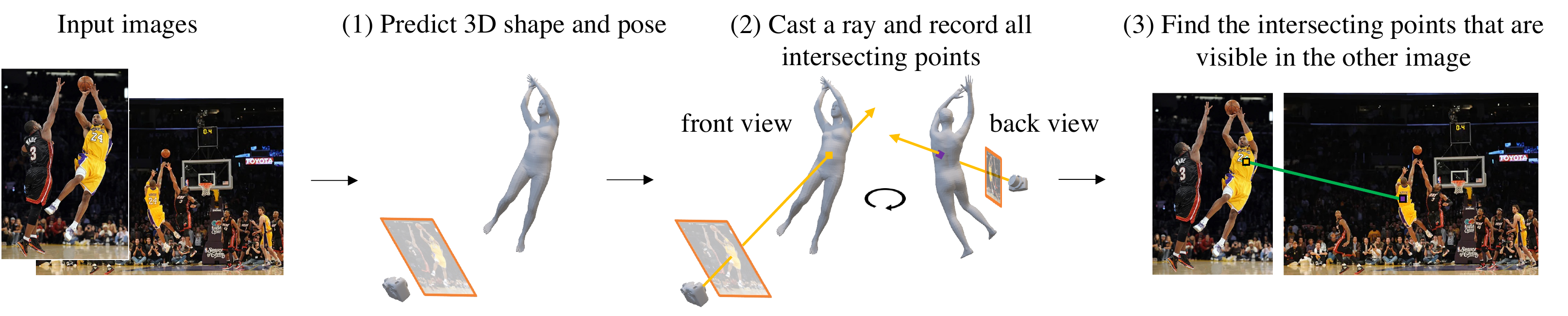}
\vspace{-2mm}
\caption{\textbf{Pipeline.} We first predict the 3D shape and pose of the basketball player from the left image. Then we cast a ray and record all the points it hits, \ie the belly button and his back. While the two images barely overlap, the right image does observe the back of player. We can thus tell that the rays of the two pixels intersect at 3D and are virtual correspondences. We conduct the same process for the right image too.}
\label{fig:vc-pipeline}
\vspace{-6mm}
\end{figure*}

\vspace{-4mm}
\paragraph{Extreme pose estimation:} There has been a surge of interest in estimating relative 3D poses among a set of little- or non-overlapping RGB(D) images \cite{yang2019extreme,qian2020associative3d,cai2021extreme,jin2021planar,shabani2021extreme}. 
Different from the classical small- or wide-baseline setup, 
the large viewpoint variations in this task result in very few co-visible regions,
rendering traditional matching-based approaches unsuitable. To address this challenge, researchers have proposed to either directly predict the transformation with deep neural nets \cite{cai2021extreme,chen2021wide}, or 
\wc{adopt the hallucinate-then-match paradigm}
\cite{yang2019extreme,yang2020extreme,qian2020associative3d,banani2020novel,germain2021visual}. 
Our work lies under the broad umbrella of the hallucination paradigm, as we derive virtual correspondences from hallucinated human shape priors and combine them with epipolar geometry. 
We adopt a pixel-level correspondence representation, which seamlessly integrates with prevailing 3D reconstruction algorithms and  can be naturally extended to the multi-view setup. In contrast, previous methods only consider two frames at a time \cite{yang2019extreme,yang2020extreme,qian2020associative3d,jin2021planar}, as 
the customized matching and optimization step prohibits them from scaling up easily.

\vspace{-3mm}
\paragraph{Structure from motion (\sfm):} Given a set of images, the goal of \sfm algorithms \cite{beardsley19963d,szeliski1994recovering,cooper1996theory,fitzgibbon1998automatic,torr1999feature,triggs1999bundle,pollefeys2004visual,agarwal2011building} is to recover both the camera poses and the (sparse) 3D geometry of the scene. Prevailing \sfm systems \cite{schonberger2016structure,snavely2006photo,snavely2008modeling,wu2011visualsfm} have enjoyed great success when the images are captured densely {with large overlapping regions}, yet they suffer drastically when the input views are sparse and have little overlap. %
To alleviate this issue, researchers have sought to exploit motion patterns \cite{angst2009static,angst2013multilinear,sweeney2019structure} or semantic keypoints of the objects \cite{dong2020imocap,xu2021widebaseline} to aid the reconstruction. However, %
they require sequences of frames as input (with static cameras) or the same set of keypoints to be visible across all views, which largely limits their applicability. Our virtual correspondences, in comparison, are much more flexible: while our VCs are also derived from objects, specifically humans, the corresponding pixels can have completely different semantics and appearances. This allows us to establish matches even if the input images have no co-visible 3D points. %
Our approach also shares similar insights with non-rigid \sfm algorithms, which  leverage shape dictionaries (\ie, priors) to constrain the solution space \cite{bregler2000recovering,del2006non,akhter2008nonrigid,del2008factorization,kong2019deepnrsfm,jensen2021benchmark,wang2020deep}.
However, unlike these approaches, we do not require 2D correspondences to be given a priori.
We instead exploit shape priors to establish VCs across views that conventionally do not have correspondences. 
As we will show in the experimental section, VCs open the door to a range of possibilities and broaden the applicable domain of \sfm.

\vspace{-3mm}
\paragraph{3D human estimation:}
Our work is also related to 3D human reconstruction approaches \cite{habermann2020deepcap,yang2020recovering,yang2021s3}. With the rise of deep learning, these methods have made tremendous progress, either from a single image \cite{kanazawa2018end,kolotouros2019learning,joo2020exemplar} or multi-view images \cite{pavlakos2020human,dong2020imocap,dong2021fast,fang2021reconstructing}.
{While these approaches mostly focus on the quality of the reconstructed shape, we attempt to recover accurate camera poses with human shape priors.}
More recently, researchers have exploited human keypoints to refine camera poses \cite{puwein2014joint,dong2020imocap}, but by virtue of VCs, our method is more flexible and does not require the same keypoints to be co-visible across views. {As we will show in Sec. \ref{sec:generalized-ba}, our bundle adjustment formulation is a superset of theirs.} 
Our work also shares similar insights with human silhouette matching \cite{sinha2004camera,sinha2010camera}, since we both do not rely on appearance matching to establish correspondences, allowing us to generalize to extreme-view setting. However, there exist several differences: First, while they require video sequences to constrain the solution space, a single image pair suffice for us. 
Second, they capitalize on sufficient motion of the object over the space for matching, whereas we exploit deep shape priors to estimate the correspondences. 
Third, their frontier points are still co-visible across cameras, yet our VCs may correspond to completely different 3D points.

\begin{table*}[t!]
    \centering

    \begin{minipage}{0.56\linewidth}
        \centering
        \includegraphics[trim={0, 0cm, 2.2cm, 3.6cm},clip,width=0.98\linewidth]{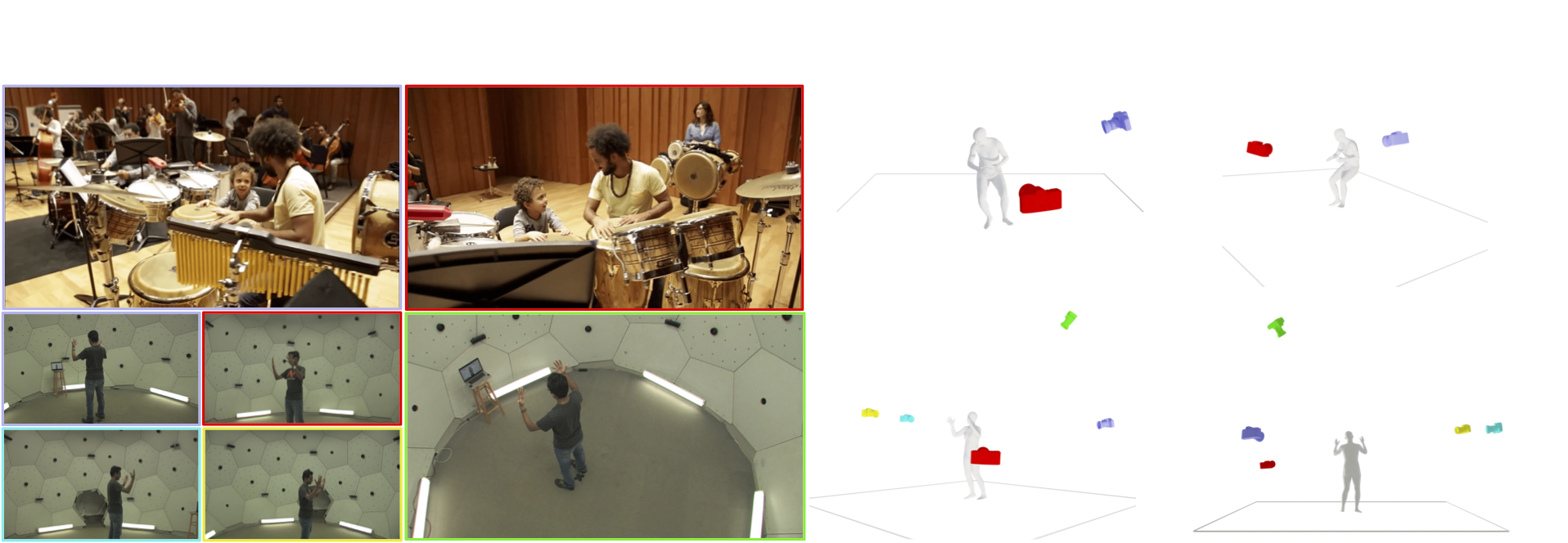}
        \vspace{-2mm}
        \captionof{figure}{\textbf{Qualitative results.} (Left) Input images. (Right) Recovered camera poses. Human meshes are for illustration purposes.}
        \vspace{-4mm}
        \label{fig:qual}       
    \end{minipage}
    \hspace{8px}    
    \begin{minipage}{0.4\linewidth}
        \centering
        \includegraphics[trim={0, 0.0cm, 0, 0cm},clip,width=0.9\linewidth]{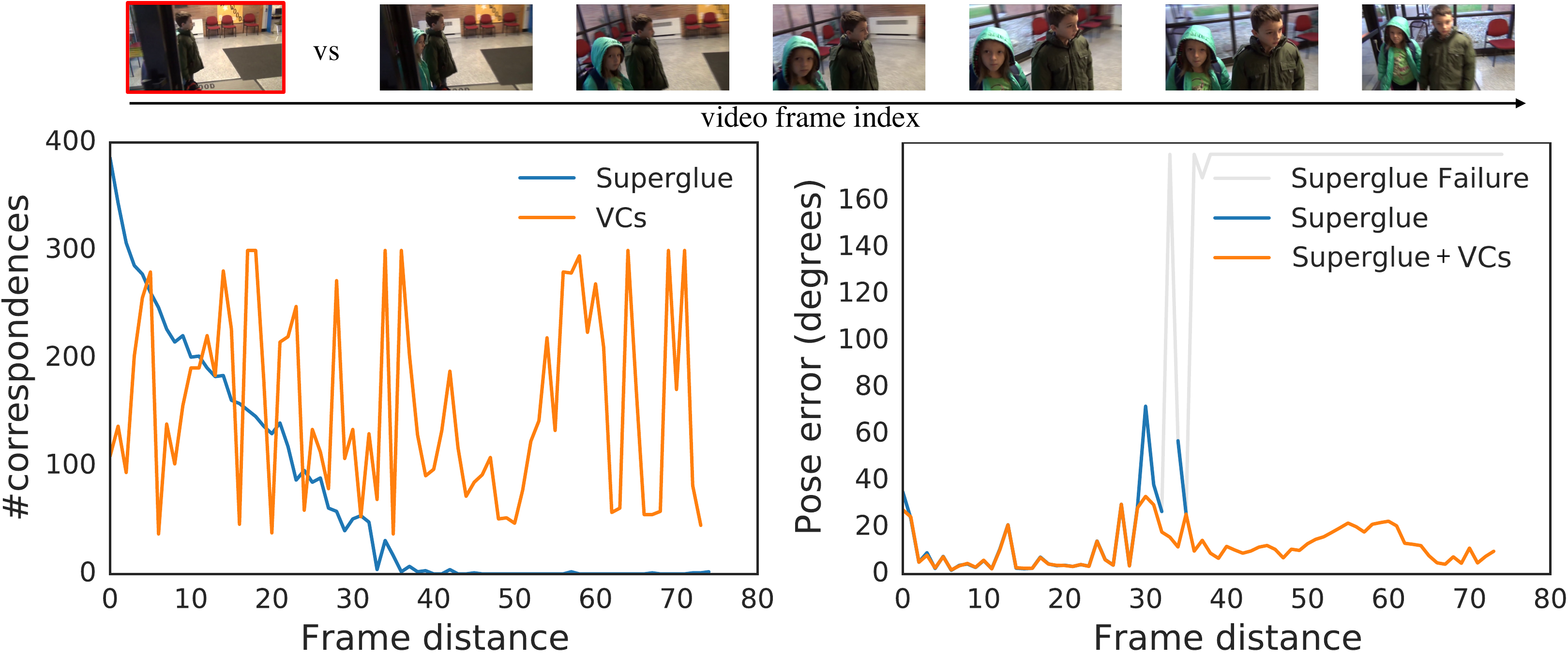} 
        \vspace{-3mm}
        \captionof{figure}{\textbf{{Effects of camera distance.}} We show the number of correspondences (left) and pose error (right) with increasing camera baseline.}
        \vspace{-4mm}
        \label{fig:transition}        
    \end{minipage}    
    \vspace{-4pt}
\end{table*}

\section{Approach}
\label{sec:method}
Our aim is to equip existing 3D systems with the ability to reason and associate images geometrically even if they have little or no overlap.
We seek to devise a method that can be seamlessly integrated with existing \shenlong{3D reconstruction frameworks} %
such that the new model can be applied to both the conventional setup and the extreme setting.
Towards this goal, we introduce a novel concept dubbed as \emph{virtual correspondence (VC)}. VCs refer to a pair of pixels whose camera rays intersect in 3D. However, unlike classic correspondences, 
they do not need to describe the same 3D points, and can have completely different semantics and appearances.  
This makes VCs much more flexible and allows VCs to be established even when 
there is little overlap among images. Importantly, VCs conform to epipolar geometry and can be combined with prevailing 3D systems naturally.
We unfold this section by formally defining VCs and discussing their relationships with existing correspondences. %
Then we present a method to estimate VCs through the lens of human shape priors. Finally we incorporate VCs into current \joycey{\sfm} formulations, %
resulting in a framework that is much more general. For simplicity, we assume there are only two cameras, but the concepts and the method can be trivially extended to the multi-camera setup (as shown in Sec. \ref{sec:exp}).

\subsection{Virtual Correspondences (VCs)}
\label{sec:vc}
We first define virtual correspondences.
Let $\cI_1$, $\cI_2\in\mathbb{R}^{H\times W \times 3}$ be the images of the same scene captured at different viewpoints and $\bp_{1}$, $\bp_{2} \in \mathbb{R}^{2}$ be the points in their respective image coordinates. 
Let $\bK_1, \bK_2 \in \mathbb{R}^{3\times 3}$ be the camera intrinsics and $[\bR_1, \bt_1], [\bR_2, \bt_2] \in \mathbb{R}^{3\times 4}$ be their extrinsic matrices. 
The ray marching from the camera center $\bo\in\mathbb{R}^3$ through $\bp$ can be written as $\br_{\bp}(d) = \bR^T(d\bK^{-1}\bar{\bp} - \bt)$, where $d > 0$ indicates the depth along the ray and $\bar{\cdot}$ refers to the homogeneous coordinate.

We say a point $\bp_{1}$ in the first image and a point $\bp_{2}$ in the second image, $(\bp_{1}, \bp_{2})$, are \emph{virtual correspondences} {if} there exists a pair of $d$'s such that:
\begin{align}
\br_{\bp_{1}}(d_{1}) = \br_{\bp_{2}}(d_{2}).
\label{eq:intersect}
\end{align}
Since there is no constraint on where the intersection should happen, the rays can intersect at (i) co-visible 3D points, (ii) 3D points that are only visible in one image (and occluded in other other), or even (iii) invisible points (\eg, free space, occupancy space, or points from occluded scene/objects).

The first scenario is exactly the definition of classic correspondences \cite{superpoint,superglue}. 
The third scenario covers many cases in semantic correspondence where the target 3D points is invisible. For instance, researchers have exploited 2D human keypoints to reconstruct 3D joints \cite{dong2020imocap,chen2021wide}. 3D joints, strictly speaking, lie within the human body and are not visible in images.
VCs can therefore be seen as a generalization of multiple \joycey{types of} existing correspondences.

In the second and third scenario, \joycey{VCs correspond to different 3D points in the scene}. %
VCs can thus have different appearances and semantics, and even describe completely different parts of the scene. We show an example in Fig. \ref{fig:vc-bunny} (right) where the pixels in the left image observe the leg while their VCs in the right see the back \joycey{of the bunny}. \cbd{We refer the readers to supp. material for more illustrations.}

Another key property of VCs is that they conform to epipolar constraints --- the two intersecting rays form an epipolar plane on which the VCs and camera origins lie.
This allows us to exploit classic geometric algorithms to establish connections among non-overlapping images, greatly expanding the applicable domains of existing 3D algorithms.
For instance, we cannot employ the five-point algorithm~\cite{longuet1981computer} for non-overlapping images in the past, since no correspondences exist. VCs, however, are more flexible and are not restricted to describing the same co-visible scene points. We can thus estimate VCs among the images and then solve for the essential matrix. 
We refer the readers to supp. material for more discussion on VCs and epipolar geometry.

While VCs are powerful, estimating them purely from 2D images is far from trivial. Without knowing the relative camera poses, one cannot exploit Eq. \ref{eq:intersect} to verify if two camera rays intersect. Furthermore, VCs may have completely different appearances and semantics, prohibiting us from employing similar approaches as classic correspondence estimators. 
Fortunately, there are many objects in the scene whose shapes we are familiar with. With such prior knowledge, we can hallucinate the shape of an object and estimate which part of the object a ray would intersect with on the other side. All one needs to do is then to find the rays (pixels) in other images that hit (see) the same intersecting point.

\subsection{Exploiting Humans for VC Estimation}

Based on the intuition above, we propose {an approach} to exploit shape priors for virtual correspondence estimation. 
We focus on humans, the most common ``objects'' in images.

Given a 2D image, we first exploit a deep network \cite{joo2020exemplar} to predict the 3D shape and pose of each person in the scene, as well as their relative poses to the camera. We use SMPL \cite{SMPL:2015} as our representation since it allows us to reconstruct a complete human mesh from partial observations.
Then we cast a ray through each pixel and record all the 3D points where the rays intersect with the human mesh via ray-plane intersection (see Fig.\ref{fig:vc-pipeline}-mid).
Finally, we identify if any of those 3D points are visible in other images by 2D-3D association. If there is, we {say} the two pixel rays intersect in 3D and the corresponding two pixels are VCs. Specifically, we use DensePose \cite{guler2018densepose} to associate each pixel with each point on the human mesh. 
{If a ray hits the back of the mesh and DensePose tells us a pixel corresponds to the back, then these two pixels are VCs.} 
Fig. \ref{fig:vc-pipeline} illustrates the process, which we repeat for all images.
We note that our formulation is generic and can be potentially applied to other objects so long as there exist proper shape priors and surface mapping. 
We show an example on cars in Sec. \ref{sec:exp}.

\begin{figure}[t]
\vspace{-3mm}
\centering
\setlength\tabcolsep{4pt}
\def\arraystretch{0.3}
\setlength{\tabcolsep}{0.5pt}
\resizebox{\linewidth}{!}{
\begin{tabular}{ccc}
\raisebox{0.7cm}{\multirow{2}{*}{\includegraphics[height=1.6cm,width=0.02\linewidth,trim={1mm, 0cm 2mm 0cm},clip]{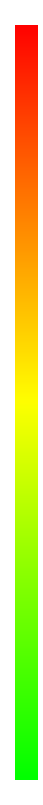}}}
&\includegraphics[width=0.31\linewidth]{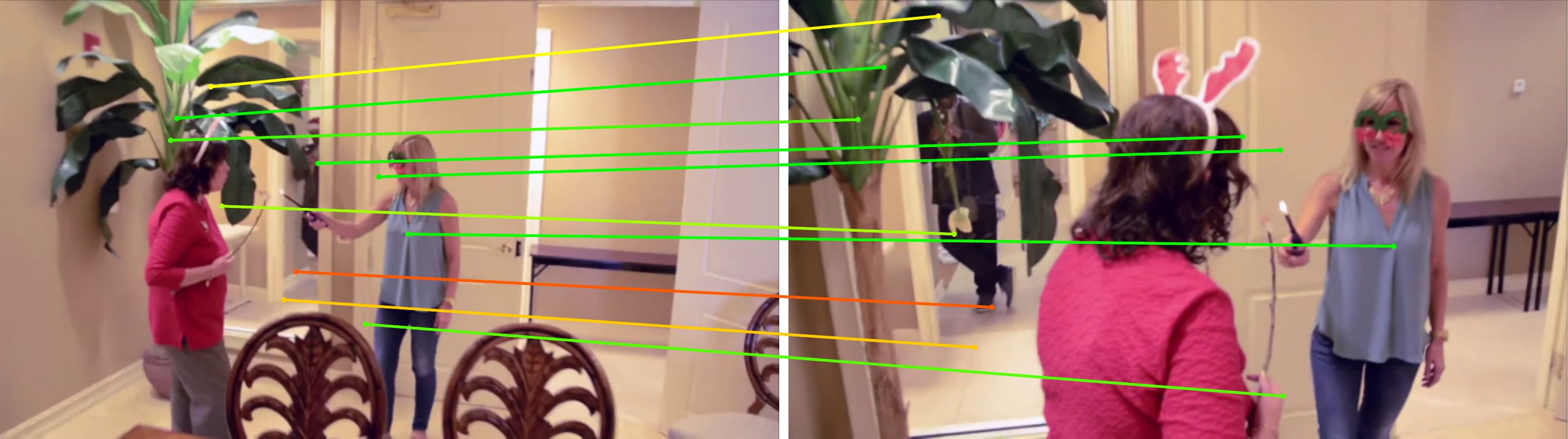}
&\includegraphics[width=0.31\linewidth]{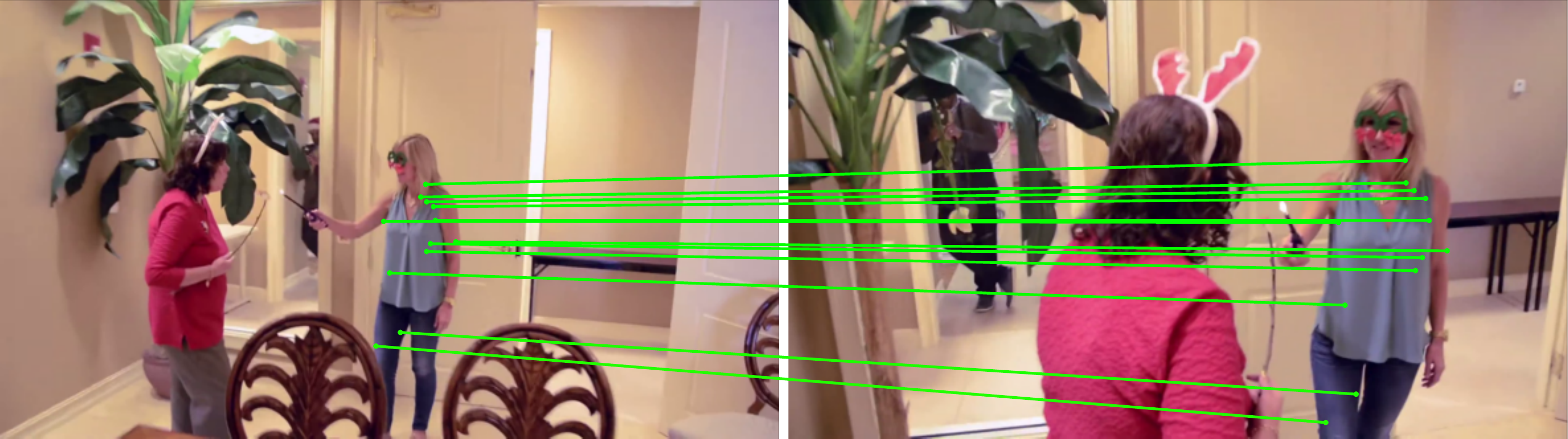}\\
&\includegraphics[width=0.31\linewidth]{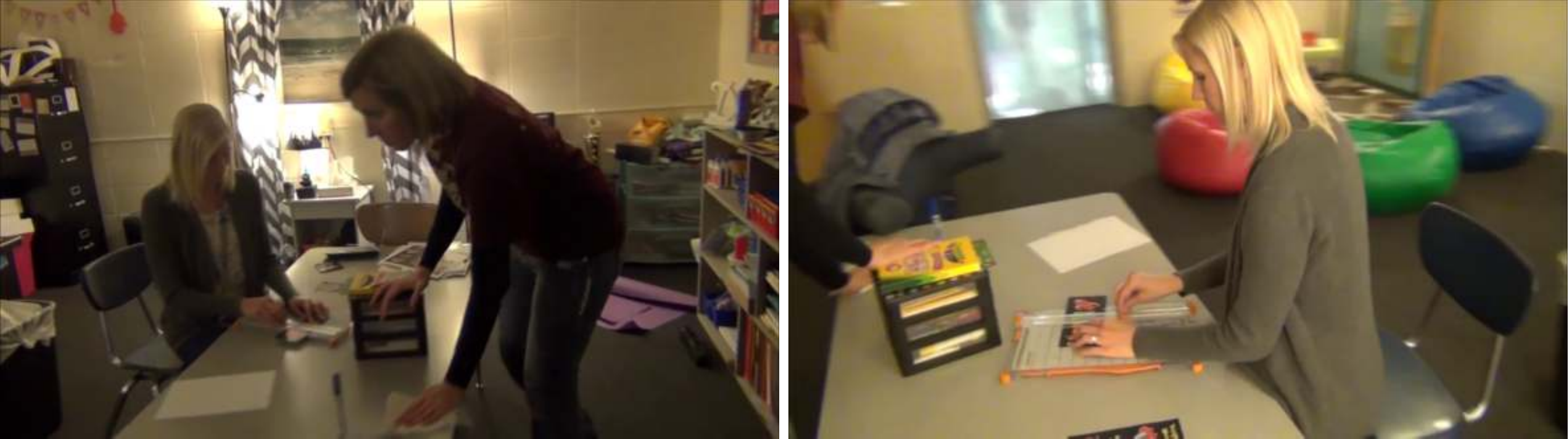}
&\includegraphics[width=0.31\linewidth]{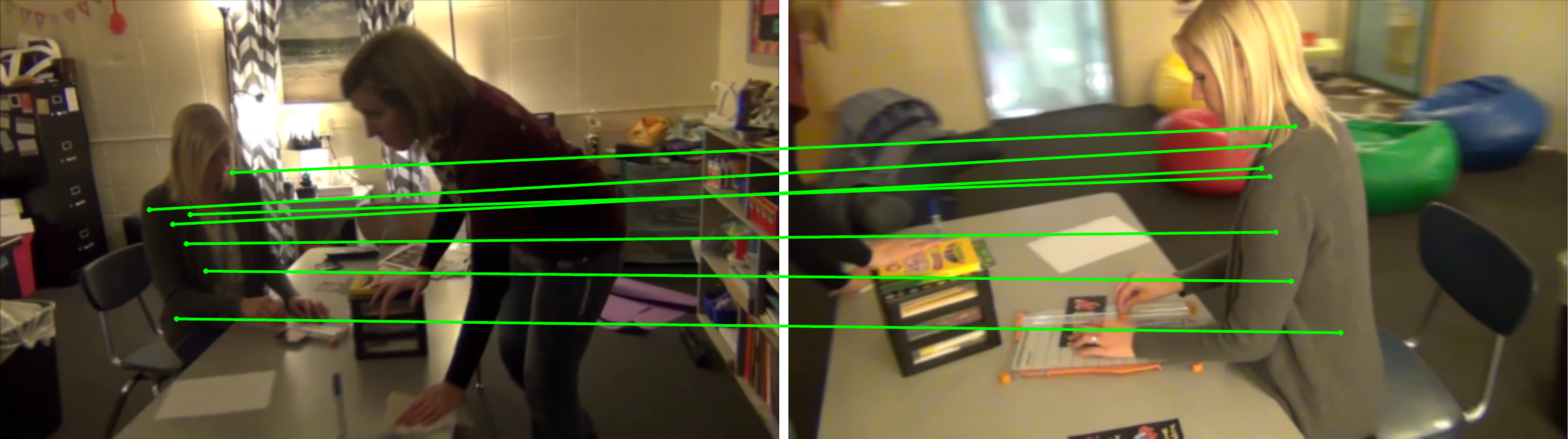}\\
&{\tiny (a) SuperGlue} & {\tiny (b) VCs}
\end{tabular}
}
\vspace{-3mm}
\caption{\textbf{Qualitative comparison.} 
Classic correspondence estimators fail when images have little overlap, since there are no co-visible 3D points. VCs can be found in both scenarios so long as the camera rays intersect.
The color indicates epipolar error.}
\vspace{-4mm}
\label{fig:vc-qual-comp}
\end{figure}

\subsection{Generalized Bundle Adjustment (BA)}
\label{sec:generalized-ba}
{Once we establish virtual correspondences, the next step is to jointly \joycey{refine} %
the camera poses as well as the sparse 3D scene geometry. Similar to classic \sfm, we initialize the camera poses using RANSAC with the five-point algorithm~\citep{hartleymultiple} in the loop\footnote{We assume the intrinsics are known or already estimated.}. But instead of employing classic correspondences, we use VCs.

Since VCs could correspond to different 3D points (see Fig.~\ref{fig:vc-bunny}), traditional triangulation approach cannot recover both 3D points.
We thus leverage the initial shape estimation (predicted by deep nets) to compute the ray-surface intersection and record the first hits for each VC. The 3D points are then registered into the global coordinate system using the estimated camera poses from the five-point algorithm.

Since the estimated structure (\ie the sparse 3D points) and poses depend heavily on the predicted shape priors, they may be noisy. We further refine the estimates by minimizing the distance between reprojected points and VCs. Formally, let $(\bX^{j_1}, \bX^{j_2})$ be the $j$-th pair of reconstructed 3D points and $(\bp_{i_1}, \bp_{i_2})$ be the associated VC pair from camera $i_1$ and camera $i_2$. Denote $\alpha = (i_1, i_2, j_1, j_2)$ as a tuple of corresponding indices. Our goal is to minimize:
\vspace{-1mm}
\begin{align}
\small
\begin{split}
\min_{\bR_i, \bt_i, \bX^{j_1}, \bX^{j_2}} &\sum_{\alpha} \lVert \bp_{i_1} - \pi_{i_1}(\bX^{j_1}) \rVert^2  + \lVert \bp_{i_2} - \pi_{i_2}(\bX^{j_2}) \rVert^2 \\
\textrm{s.t.} \quad \Big((&\bX^{j_1} - \bo_{i_1}) \times (\bX^{j_2} - \bo_{i_2})\Big)^T (\bo_{i_2} - \bo_{i_1}) = 0, %
\label{eq:generalized-ba}
\end{split}
\end{align}
\vspace{-0mm}
where $\pi_{i}(\bX) \sim \mathbf{K}_i(\mathbf{R}_{i} \bX + \mathbf{t}_{i})$ is the perspective projection operator, 
and the constraint enforces the two camera rays to be co-planar such that epi-polar geometry holds.

Using the constraint, we can further re-write one VC point as a function of the other:
\begin{align}
\begin{split}
\bX^{j_2} = \bX^{j_1} + a^j \cdot (\bX^{j_1} - \bo_{i_1}) + b^j \cdot (\bo_{i_2} - \bo_{i_1}). \\
\end{split}
\label{eq:reparam}
\end{align}
The two free parameters $a^j$ and $b^j$ can be thought of as the ``thickness'' of the shape between the intersecting points. When both parameters become 0, the two 3D points merge into one, and VCs reduce to classic correspondences. 

By replacing Eq. \ref{eq:reparam} into Eq. \ref{eq:generalized-ba}, we obtain an unconstrained minimization problem that is similar to, yet more generic than classic BA. Instead of refining a set of \emph{co-visible 3D points}, we now adjust a bundle of \emph{point tuples}. 
We, however, note that classic correspondences extracted with conventional methods such as SuperGlue \cite{superglue} can still fit into this formulation by fixing $a^j = b^j = 0$.
We use L-BFGS~\cite{nocedal1980updating} to solve this non-linear least square problem. In practice, we treat Eq. \ref{eq:reparam} as a soft constraint  since it works slightly better. 
We refer the readers to supp. material for more discussions.

\begin{figure}
\vspace{-3mm}
\setlength\tabcolsep{4pt}
\def\arraystretch{1.05}
\resizebox{\linewidth}{!}{
    \begin{tabular}{cc}
\includegraphics[trim={0, 0cm, 0, 1cm},clip,width=0.48\linewidth]{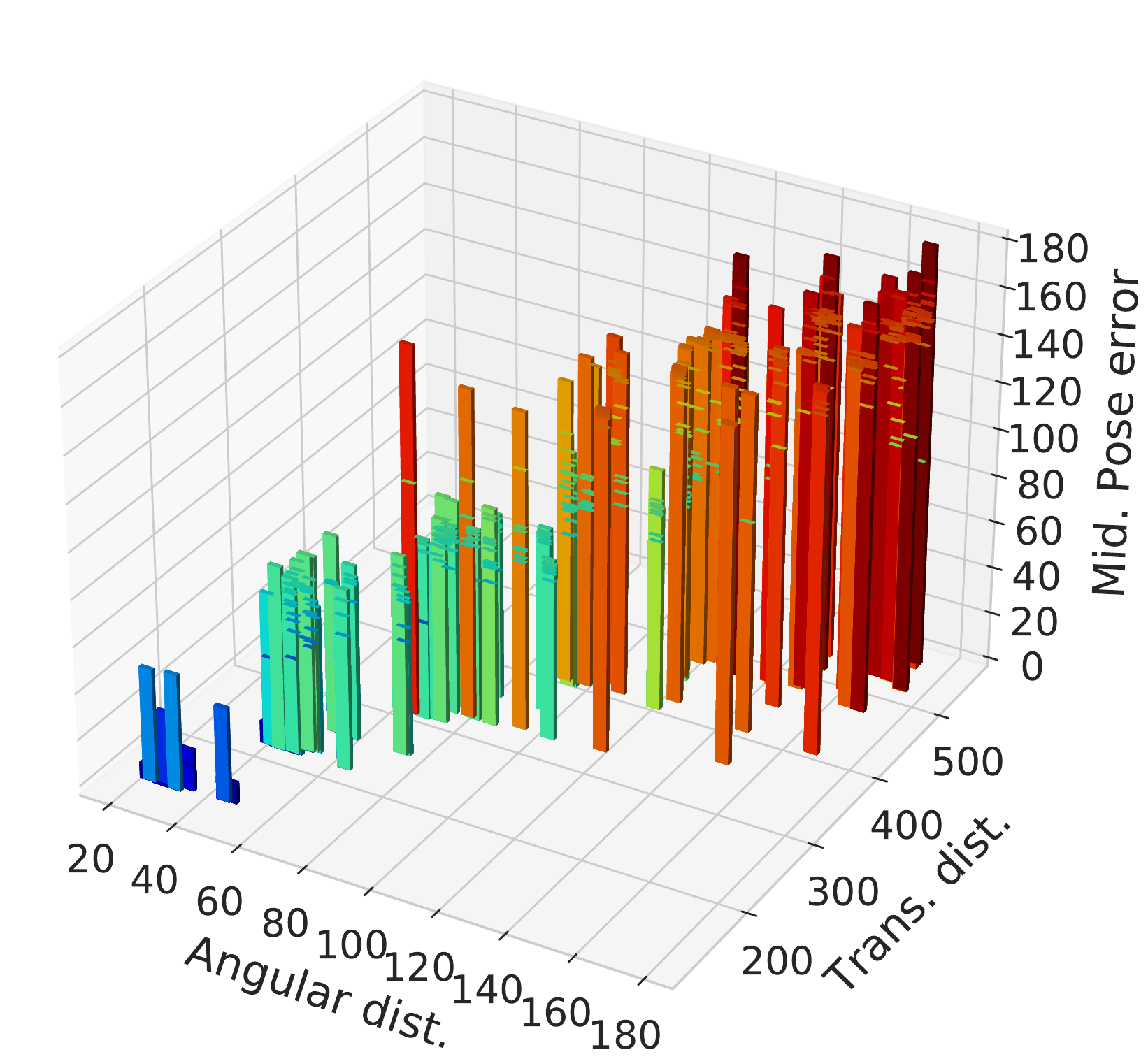}
& \includegraphics[trim={0, 0cm, 0, 1cm},clip,width=0.48\linewidth]{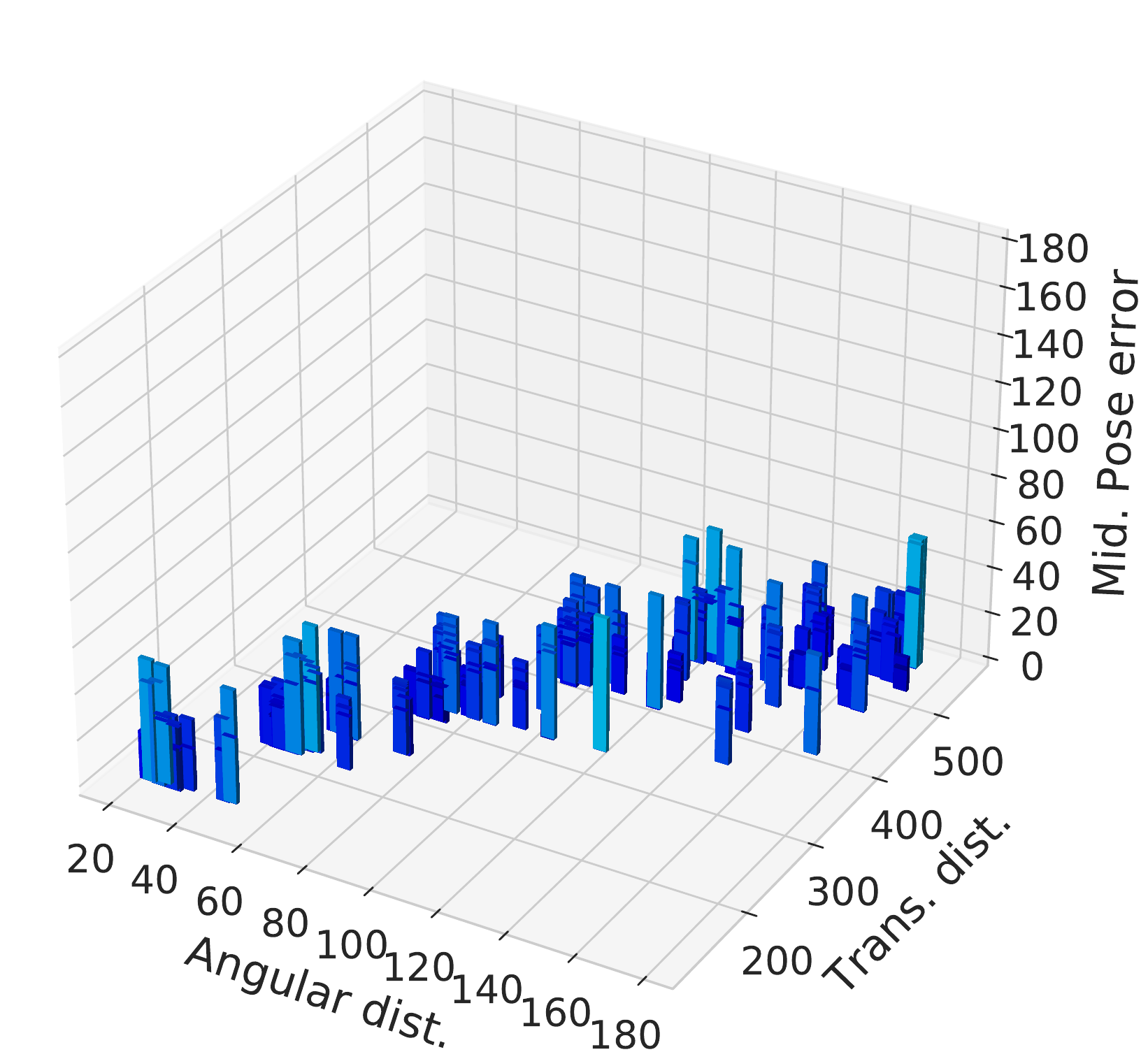}\\
(a) Classic \sfm & (b) Our \sfm
    \end{tabular}
}
\vspace{-2mm}
\caption{{\bf Pose error vs. ground-truth pose distance.} The median pose error in classic \sfm (left) increases with increasing camera baseline, while the median pose error for our method (right) stays low regardless of viewpoint differences.
}
\vspace{-4mm}
\label{fig:3dbar}  
\end{figure}

\paragraph{Discussion:}
{VCs can be combined with classic correspondences to improve the overall robustness and performance of 3D reconstruction systems (see Sec. \ref{sec:exp}). When the images barely overlap and few classic correspondences are available, the system can rely on VCs to recover the world and camera geometry. When the images do overlap, VCs can serve as additional visual cues and regularizers. VCs thus significantly expand the applicable setting of existing SfM systems.}

\vspace{-2mm}
\section{Experiments}
\vspace{-2mm}
\label{sec:exp}

{In this section, we first evaluate the effectiveness of virtual correspondence and our 3D system on two challenging datasets. Then we comprehensively study the characteristics of our method. With the estimated camera poses, we further conduct two downstream tasks, namely scene reconstruction with multi-view stereo and novel view synthesis, in difficult extreme-view cases. Finally, to showcase that our method generalizes beyond human-based images, we demonstrate proof-of-concept results with cars.}
\vspace{-1mm}
\subsection{Datasets} 
\vspace{-1mm}
\paragraph{CMU Panoptic dataset:} CMU Panoptic dataset~\citep{Joo_2015_ICCV,Joo_2017_TPAMI} is a large-scale, multi-view video dataset designed for human analysis. It provides {ground-truth} camera poses as well as person associations across views. \wc{The sequences were captured {in a studio with} (approximately) synchronized cameras widely spread across the dome, providing us a diverse set of viewpoints that are barely available in the real world (\eg, cameras looking at a person from the top)}. We select 43 sequences from \texttt{pose}, \texttt{haggling}, and \texttt{dancing}. Each sequence contains 1$\sim$3 people performing different actions. We divide the data into two splits. Each split comprises a set of unique sequences and cameras without any overlap. Due to image quality, we only consider the videos captured by HD cameras. We sample a frame every five seconds to avoid similar human poses. We also run human detection on each sampled frame. If no person is present in the scene, we discard the frame. In total, we obtain 2955 image sets for each split, with each set containing 15-16 camera views. 
We refer the readers to the supp.~material for more details.

\paragraph{Mannequin Challenge:} Mannequin Challenge (MC)~\citep{li2019learning} is a dataset of {internet} video clips where the participants stay still in different poses, while the
video-takers 
move freely in space and capture the event. 
These videos, by design, allow us to look at {a static} scene from various angles.
We follow a similar pipeline as \citep{li2019learning} to reconstruct the \joycey{ground-truth} camera trajectories and filter out snippets {with small shifts in viewpoints or view directions}. %
In the end, we obtain 18 video snippets where the cameras rotate by at least 90$^\circ$ within each sequence. To further increase pose diversity, we additionally collect 6 MC videos ourselves. Compared with the CMU dataset, the camera poses in MC videos are rather \emph{generic} \cite{freeman1994generic}, yet the {background} scenes, which consist of both indoor and outdoor environments, are much more diverse. 
Finally, for each snippet, we compute the pose difference between each frame and the first frame. We sample a frame at every 20 percentile of the snippet and obtain $\sim$200 image pairs. All the images are treated as the test set.

\begin{table}
\vspace{-3mm}
\centering
\setlength\tabcolsep{4pt}
\def\arraystretch{1.05}
\scalebox{0.7}{
\begin{tabular}{lcccccc}
\specialrule{.2em}{.1em}{.1em}
Pose estimation AUC ($\uparrow$) & \multicolumn{3}{c}{CMU Panoptic Studio} & \multicolumn{3}{c}{Mannequin Challenge}\\
Methods & @$15^\circ$ & @$30^\circ$ & @$45^\circ$  & @$15^\circ$ & @$30^\circ$ & @$45^\circ$ \\
\hline
SuperGlue \cite{superglue}  & 10.02 & 16.74 & 19.36 & 26.38 & 34.85 & 39.10  \\
LoFTR \cite{sun2021loftr} &5.12 &10.47 &13.07 &27.47 & 35.98 & 40.10\\
SIFT \cite{sift} + BA \cite{schonberger2016structure} & 7.68 & 11.39 & 13.33  & 14.17 & 20.24 & 24.25 \\
SuperPoint \cite{superpoint} + BA \cite{schonberger2016structure} & 9.22 & 13.77 & 15.85   & 17.12 & 23.48 & 26.81 \\
SuperGlue \cite{superglue} + BA \cite{schonberger2016structure} & 10.68 & 16.57 & 18.92  & 26.24 & 35.12 & 39.46 \\
LoFTR \cite{sun2021loftr} + BA \cite{schonberger2016structure} &8.35& 14.52& 17.01& 27.51& 36.32& 40.55 \\
Deep regression \cite{relnet} & 14.36 & 18.60 & 23.18 & 4.61 & 11.23 & 16.44\\
Deep optimization \cite{besl1992method,kanazawa2018learning} & 7.88 & 27.17 & 42.42 & 15.38 & 47.08 & 63.67 \\
Our \sfm & \textbf{18.21} & \textbf{46.05} & \textbf{62.08}  & \textbf{36.24} & \textbf{61.38} & \textbf{73.20}\\
\specialrule{.1em}{.05em}{.05em}
\end{tabular}
}
\vspace{-2mm}
\caption{{\bf Two frame relative pose estimation} on the CMU dataset and the MC dataset. First two rows perform five-point algorithm to derive camera poses. BA = Bundle Adjustment.
}
\vspace{-4mm}
\label{tab:cmu-mc-full}
\end{table}

\subsection{Experimental Details}

\paragraph{Metrics: } Following previous work~\cite{yi2018learning,zhang2019learning,brachmann2019neural,superglue}, we employ the area under the cumulative error curve (AUC) to evaluate the recovered camera poses. 
We report the AUC at three different thresholds ($15^\circ$, $30^\circ$, and $45^\circ$). The pose error is defined as the maximum of 1) the angular difference between predicted and GT rotation vectors; and 2) the angular difference between predicted and GT translation vectors.
{We report angular difference for translation since it can only be recovered up to a scaling factor \cite{hartleymultiple}.}
{As for 3D reconstruction, there is no standard protocol to compare point clouds directly produced by \sfm systems because each \sfm algorithm can choose which 3D points to reconstruct. Furthermore, there is no ground-truth shape for both datasets. We thus follow \cite{joo2018total} to compute the silhouette accuracy between the rendered mask and the 2D segmentation mask.}

\paragraph{Baselines: } 
We compare our method against a wide range of relative pose estimation methods.
For traditional matching-based methods, we first detect the key points and extract their corresponding features with SIFT \cite{sift} or SuperPoint \cite{superpoint}. 
We then establish classic correspondences with either nearest neighbour matching with ratio test~\cite{sift}} or SuperGlue (SG)~\cite{superglue}. We also compare with LoFTR~\cite{sun2021loftr}. We further use RANSAC \cite{fischler1981random} coupled with the five-point algorithm to filter outliers.
We then incrementally recover and bundle adjust the image poses with COLMAP \cite{schonberger2016structure}. Alternatively, if there are only two views, we also perform pose estimation with the five-point algorithm and essential matrix decomposition.
Next, for deep regression methods, we employ a state-of-the-art pose estimation network \cite{en2018rpnet} to predict the relative camera pose between an image pair. 
Finally, we compare against a deep optimization approach that estimates camera poses by aligning 3D shapes. The baseline is inspired by the state-of-the-art indoor extreme pose estimation method~\cite{qian2020associative3d} and can be seen as a variant for humans. 
Specifically, we utilize the latest EFT-Net~\cite{joo2020exemplar} to reconstruct 3D human models and align them with ICP~\cite{besl1992method}. To avoid local minima, we first register the shapes based on their canonical coordinates. Next, we associate each part of the shape based on its semantics. We further prune out the limbs and exploit only torso and head during matching since these two parts are more robust in practice. These strategies drastically improve the performance of this baseline.

\begin{table}
\vspace{-3mm}
\centering
\setlength\tabcolsep{4pt}
\def\arraystretch{1.05}
\scalebox{0.85}{
        \begin{tabular}{ccccccc}
        \specialrule{.2em}{.1em}{.1em}
        \multicolumn{2}{c}{Initialization} & \multicolumn{2}{c}{BA} & \multicolumn{3}{c}{Pose estimation AUC}  \\
        SG  & VCs & SG & VCs & @$15^\circ$ & @$30^\circ$ & @$45^\circ$\\
        \hline
        \checkmark & - & - & - & 10.02 & 16.74 & 19.36 \\ 
        \checkmark & \checkmark & - & - & 10.29 & 31.27 & 48.96 \\
        \checkmark & - & \checkmark & - & 10.68 & 16.57 & 18.92\\
        - & \checkmark & - & \checkmark & 15.89 & 43.92 & 60.38 \\
        \checkmark & \checkmark & \checkmark & \checkmark & \textbf{18.21} & \textbf{46.05} & \textbf{62.08}\\
        \specialrule{.1em}{.05em}{.05em}
        \end{tabular}
}
\vspace{-2mm}
\caption{{\bf Ablation study} on the CMU dataset. SG = SuperGlue.
}
\vspace{-4mm}
\label{tab:system-ablation}
\end{table}

\paragraph{Implementation details: }
Our 3D system considers both classic correspondences and VCs. We exploit SuperGlue \cite{superglue} to estimate classic correspondences and ReID-Net \cite{torchreid} to match a person across multiple viewpoints. 
For the deep regression baseline, we train and validate on the training split of CMU dataset. For the rest of the learning based approaches, \emph{including our method}, we adopt the pre-trained weights {provided by the authors} and conduct inference only.

\subsection{Experimental Results}

\paragraph{CMU Panoptic Studio: } 
As shown in Tab.~\ref{tab:cmu-mc-full}(left), our \sfm outperforms all baselines at all thresholds in the two-frame pose estimation task. SuperGlue \cite{superglue} ranks second when the pose error threshold is low, but deep optimization \cite{besl1992method,kanazawa2018end} surpasses it when the threshold increases. 
This is expected since matching-based approaches can produce accurate estimation when classic correspondences are available, yet fail catastrophically when the viewpoints are very different. 
Deep optimization, in contrast, is not as accurate when the view difference is small, but has fewer fatal failures.
\joycey{Our approach, which exploits both classic and virtual correspondences, does not suffer from either 
catastrophic wide-baseline failures or inaccurate narrow-baseline matching.} 
\begin{wraptable}{r}{0.36\linewidth}
    \vspace{-0mm}
    \centering
    \setlength\tabcolsep{5pt}
    \def\arraystretch{1.05}
    \resizebox{\linewidth}{!}{
        \begin{tabular}{c}
        \includegraphics[trim={0mm, 0, 0, 0},clip,width=1\linewidth]{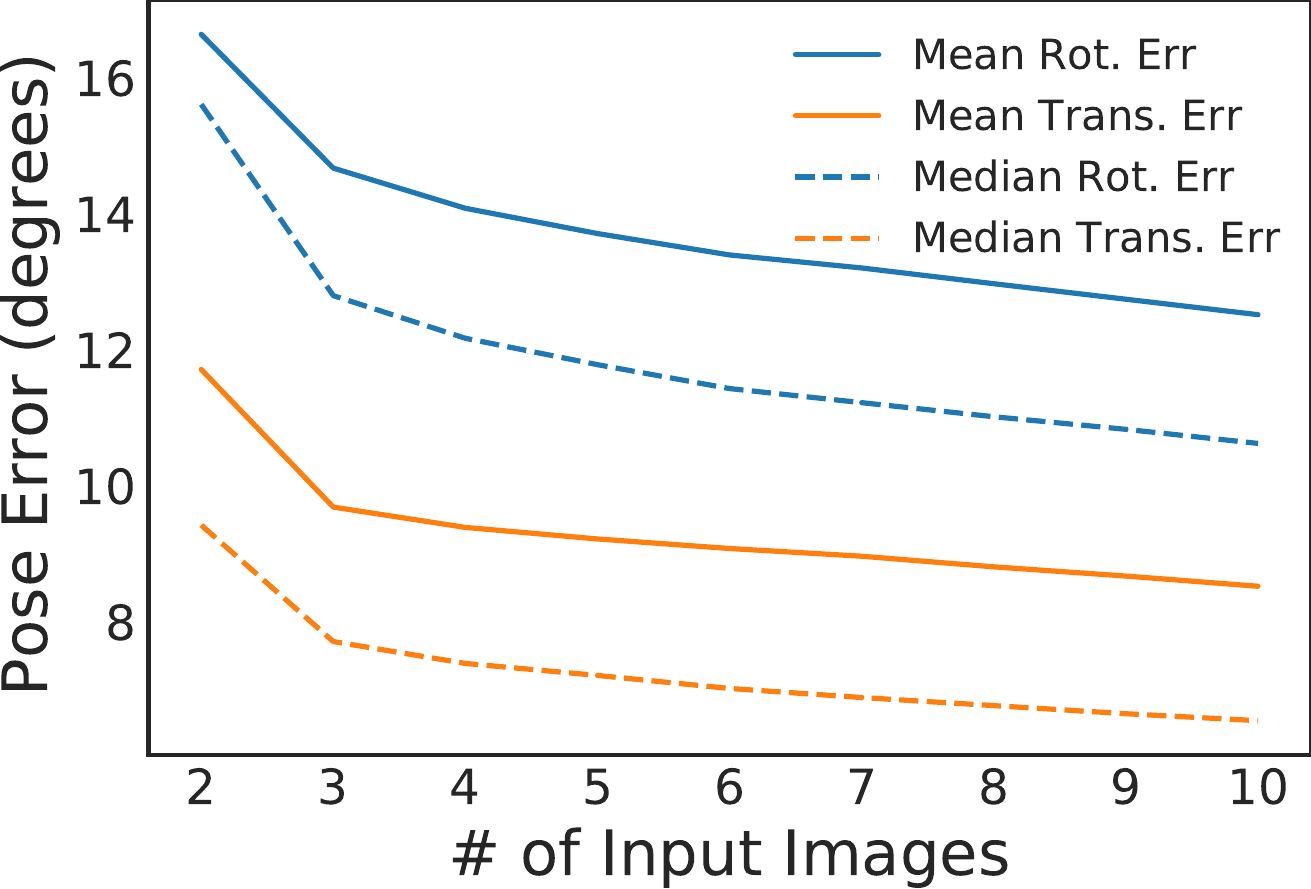}
        \end{tabular}
    }
    \vspace{-8pt} 
    \captionof{figure}{\textbf{Error vs. \# of images.}}    
    \label{table:perf_n}
    \vspace{-10pt}
\end{wraptable}

Our \sfm has a median error of $15.7^\circ$ and the pose error at the 80th percentile is less than $24^\circ$. 
In contrast, the median error of deep optimization is $23.5^\circ$ and the pose error at the 80th percentile is $44^\circ$. 
Compared to EFT-Net, we improve the silhouette accuracy from 74\% to 81\%. 

We also investigate how our \sfm scales with more input images. Following COLMAP \cite{schonberger2016structure}, we start from an image pair and then incrementally register new images. As shown in Fig.~\ref{table:perf_n}, the pose error reduces as more images are added. The reduction is most significant when registering the third image. We hypothesize this is because the third image greatly increases the overlap among the images, providing more reliable classic correspondences during bundle adjustment. We also compare our approach with the classic \sfm methods. Our AUC continuously outperforms the baselines at all thresholds (\eg, $@15^\circ$: 28.4 vs 17.6). We refer the readers to the supp.~material for full ablation table, cumulative error plots, and detailed performance of all methods with respect to the input images.

\begin{figure}
\centering
\includegraphics[trim={5cm, 8cm, 3cm, 0},clip,width=0.43\linewidth]{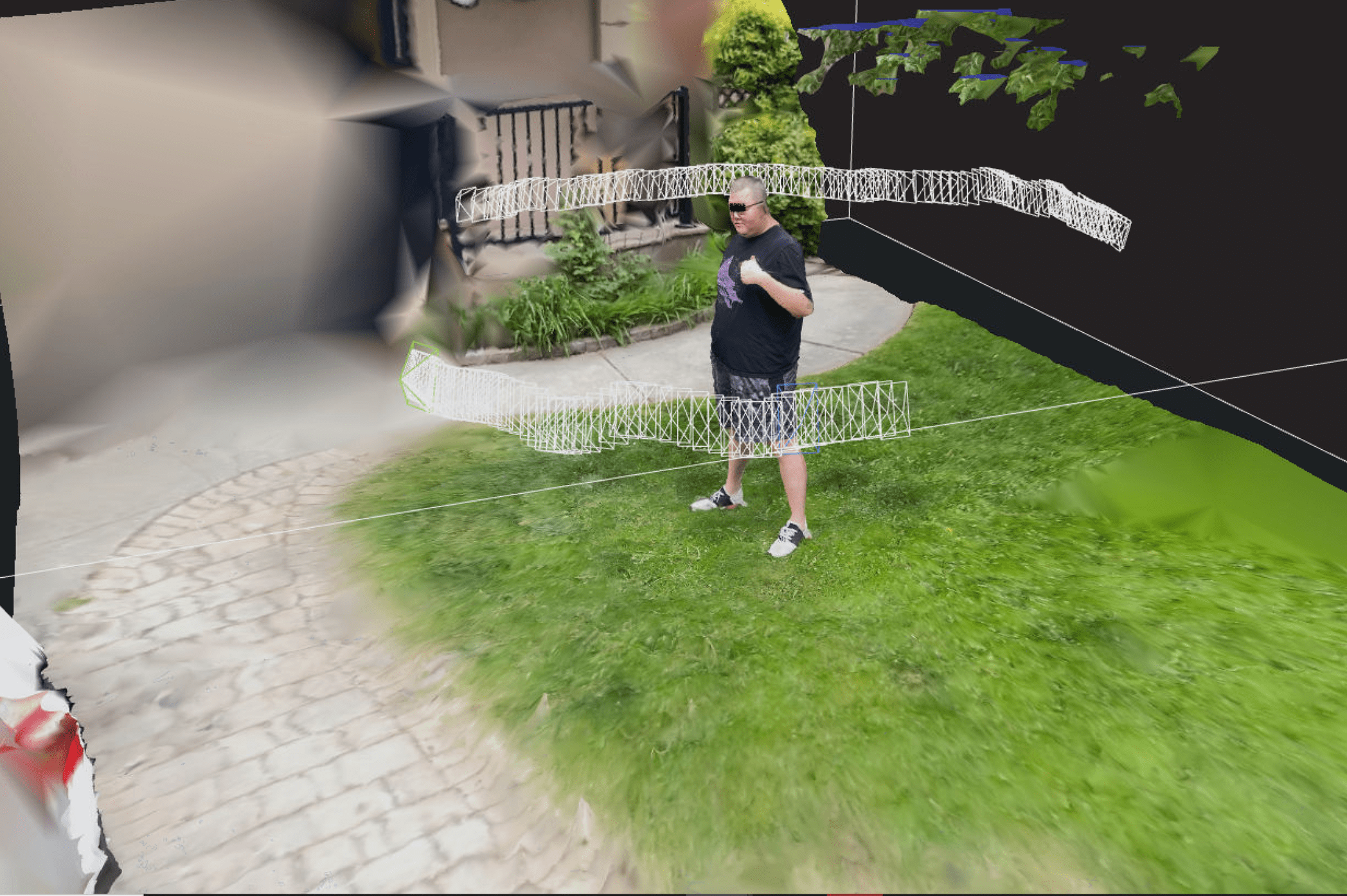}
\includegraphics[trim={1cm, 5cm, 7cm, 3cm},clip,width=0.45\linewidth]{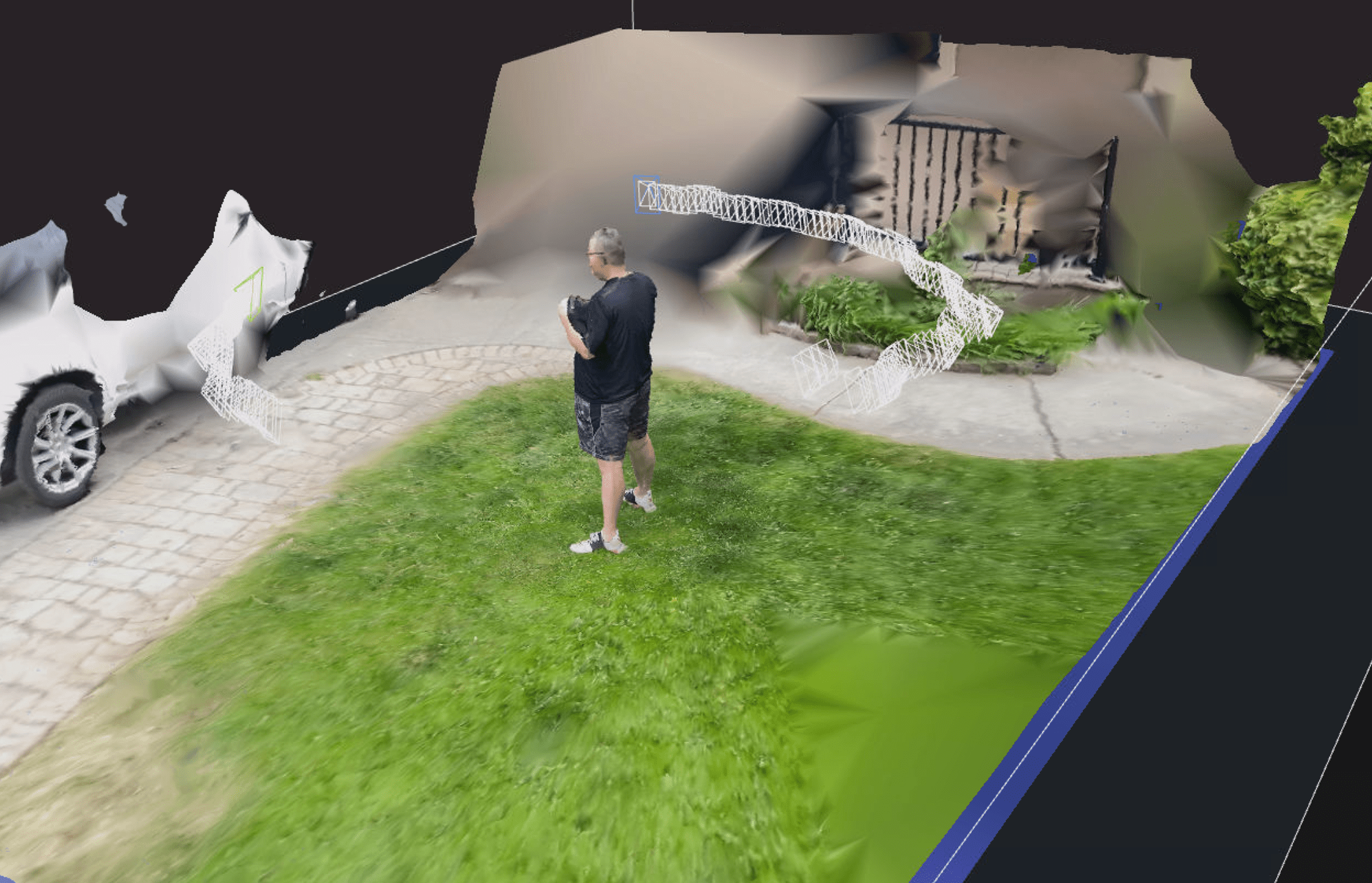}
\vspace{-2mm}
\caption{\textbf{Reconstructed mesh} using our method + multi-view stereo for two non-overlapping video sequences.}
\vspace{-4mm}
\label{fig:mvs}
\end{figure}

\vspace{-2mm}
\paragraph{Mannequin Challenge: }
As shown in Tab. \ref{tab:cmu-mc-full}(right), our method outperforms all baselines at all thresholds. 
Despite more diverse scenes, our AUC on the MC dataset is higher than that of the CMU dataset. 
We hypothesize this is because the viewpoint changes in the MC dataset are less significant than the CMU dataset, 
due to how the dataset was collected. 

\vspace{-2mm}
\paragraph{Qualitative results: } 
{We showcase our results on a two-view MC image pair, and a five-view CMU image set in Fig.~\ref{fig:qual}. Our testing scenarios are typically very challenging, with large view variations and small proportion of co-visible regions. Nevertheless, our proposed \sfm framework is able to recover both relative poses as well as the parametric human shape accurately. }

\subsection{Analysis}

\paragraph{Ablation study: } 
To gain more insights into the contribution of each component, we evaluate our method with different configurations on the CMU dataset. 
As shown in Tab. \ref{tab:system-ablation}, by simply exploiting VCs during initialization, our method surpasses classic \sfm in terms of AUC at large thresholds. 
Additionally, {the ablation study shows that} bundle adjustment is critical for VCs. 
We conjecture this is because VCs are constructed from initial shape priors, which are noisy. 
By bundle adjusting the line segments, we are essentially conducting maximum likelihood estimation \cite{hartleymultiple} under Gaussian noise assumption on VC re-projection errors, \joycey{which mitigates errors introduced by inaccurate VC pairs}.

\begin{figure}
\centering
\includegraphics[trim={0, 0cm, 0, 0},clip,width=0.9\linewidth]{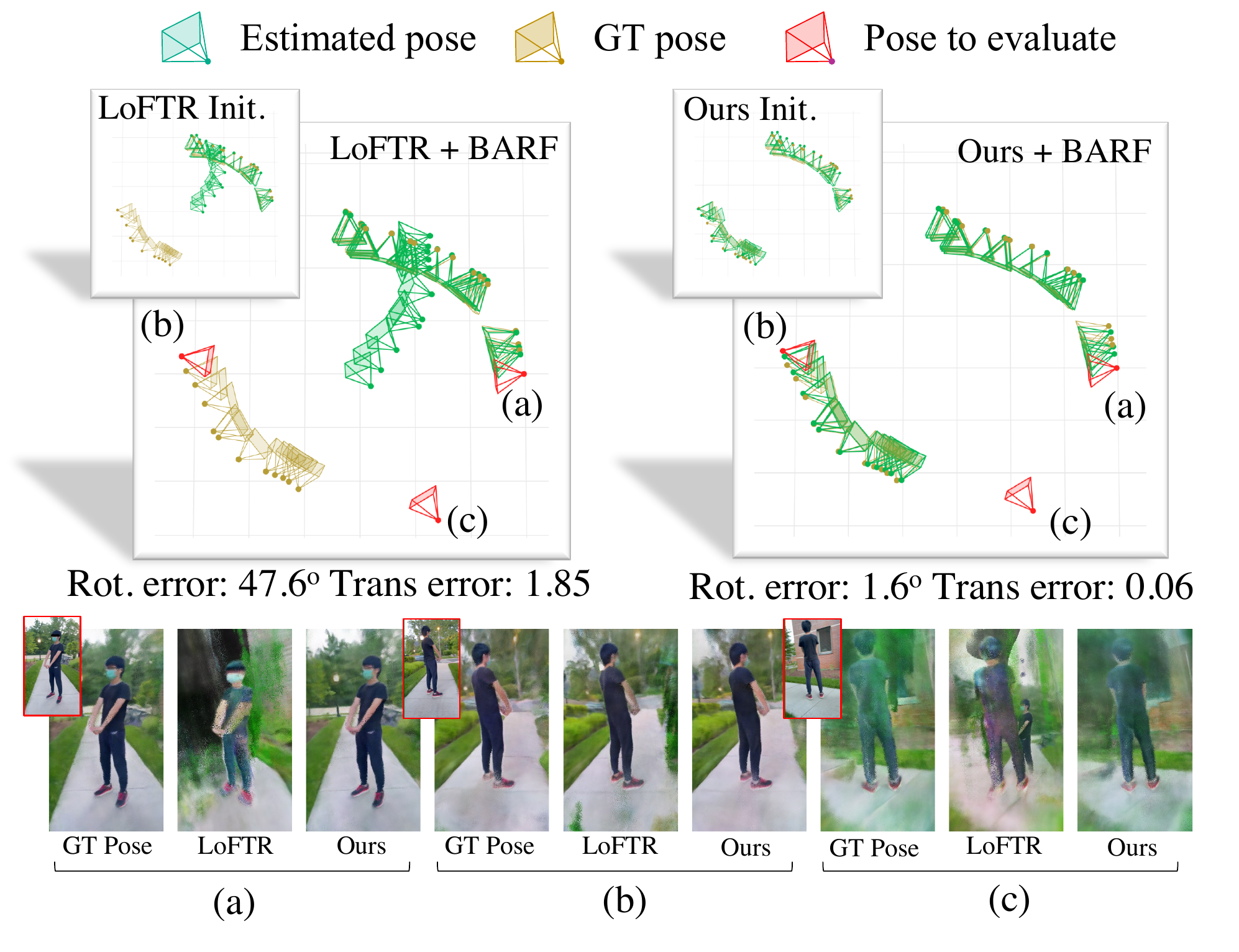}
\vspace{-4mm}
\caption{\textbf{Novel view synthesis:} (top left) camera poses initialized with LoFTR and refined by BARF; (top right) camera poses initialized with our method and refined by BARF; (bottom) images synthesized at novel views by BARF initialized with GT pose, LoFTR and our framework respectively.}%
\vspace{-4mm}
\label{fig:barf}
\end{figure}

\paragraph{Effects of viewpoint changes: } 
We use the MC dataset to illustrate how classic and virtual correspondences evolve with viewpoint changes and how it affects pose estimation. %
In general, the ground-truth camera pose difference is proportional to the video frame index distance. For each video, we compute the classic correspondences and VCs between all frames and the first frame, and then estimate the relative camera poses based on them. Since the number of classic correspondences decreases drastically when the viewpoint changes, classic \sfm fails. In contrast, our \sfm framework incorporates both classic correspondences and VCs \joycey{to avoid failures}.
Fig.~\ref{fig:transition} shows an example of how our system produces decent estimation across all distances. We also showcase a ``discrete" evaluation on the CMU dataset in Fig.~\ref{fig:3dbar}. The pose error of classic \sfm methods increases significantly with respect to the \joycey{ground-truth} camera pose distances (the diagonal direction), while our \sfm performs consistently across all settings.

\vspace{-2mm}
\paragraph{Reliability of human parts: } 
We compute the histogram over all VCs on both datasets. Around half of the VCs lie on human torso, and around 12\% of VCs are derived from the human head. The remaining VCs uniformly spread across the whole body. Unlike the deep optimization baseline, we do not encode any prior knowledge into our system, yet our approach is able to automatically discover that human torso is the most reliable parts within the predicted 3D shapes. Fig.~\ref{fig:vc-qual-comp} shows a subset of VCs selected by our \sfm system.

\vspace{-2mm}
\paragraph{Generalization to in-the-wild images: } 
Our approach can be applied to real-world image collections without bells and whistles. 
We test our system on a pair of movie frames and two pairs of sports photos in Fig. \ref{fig:teaser-new}. 
Even though the cameras are far apart and the images are slightly asynchronous, our system still produces reasonable estimates. 
More results on classic movies and sports events can be found in supp. materials.

\vspace{-2mm}
\paragraph{Limitations: }
\label{subsec:limitation}
{
Our approach relies heavily on the predicted shape priors. 
While we can handle noisy predictions by pruning out the outlier VCs during geometric verification,
if the initial estimation is completely wrong (which our system can detect by comparing silhouette consistency, DensePose consistency, \emph{etc.}), we will not be able to construct VCs.
Additionally, similar to classic \sfm algorithms, we assume the scene is static. While we can tolerate slight movements (see Fig. \ref{fig:teaser-new}), it fails when human poses change significantly. 
}

\begin{figure}
\centering
\includegraphics[trim={1.2cm, 1.5cm, 1.2cm, 0.5cm},clip,width=0.9\linewidth]{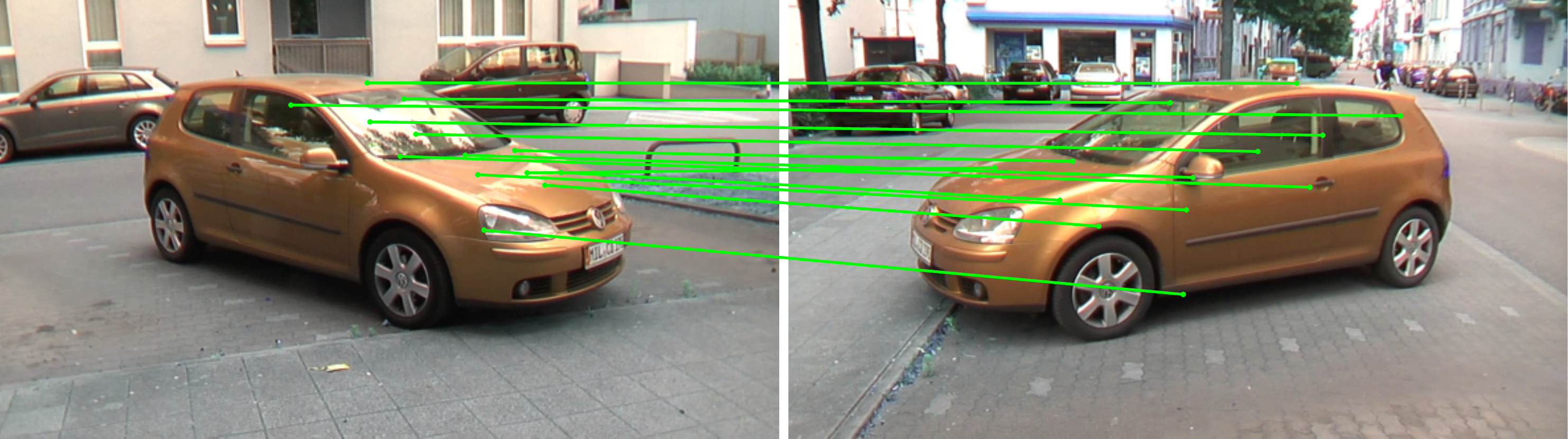}
\vspace{-3mm}
\caption{\textbf{Virtual correspondences from cars.}}
\vspace{-5mm}
\label{fig:vc-cars}
\end{figure}

\subsection{Applications}
\paragraph{Scene reconstruction with multi-view stereo (MVS):} 
\shenlong{We first show how VCs enable coherent multi-view 3D reconstruction from non-overlapping videos.}
This type of capture is common in practice, yet most \sfm and MVS systems can only handle each sequence individually, resulting in two disjoint 3D reconstructions. Our VCs, in contrast, are able to recover relative poses even from non-overlapping images, which allows us to obtain a single, coherent MVS point cloud to unlock further geometry processing such as mesh reconstruction.
We use {RealityCapture}~\cite{realitycapture} to reconstruct the 3D scene by initializing the camera poses with our estimation.
The resulting high-quality meshes suggest that the recovered camera poses and the extracted point clouds are accurate (see Fig. \ref{fig:mvs}). 
\shenlong{In contrast, both RealityCapture's built-in 3D reconstruction pipeline and COLMAP~\cite{{schonberger2016structure}} fail due to non-overlapping viewpoints. }

\vspace{-2mm}
\paragraph{Novel view synthesis: } 
We further demonstrate the effectiveness of our approach \joycey{in extreme-view scenarios} through the task of novel view synthesis, which relies heavily on input poses.
In particular, we adopt BARF \cite{barf}, an approach that can learn a neural radiance field \cite{mildenhall2020nerf} and refine camera poses simultaneously.
\joycey{We again use two non-overlapping video sequences.} 
We initialize BARF with the poses recovered by our method and LoFTR \cite{sun2021loftr} respectively. 
Our estimated poses, which are already fairly accurate, are further refined through the course of BARF training (see Fig. \ref{fig:barf}).
In contrast, LoFTR \cite{sun2021loftr} fails to estimate the relative camera poses among the two sequences correctly (see the {\color{ForestGreen} green} cameras) and \joycey{the resulting BARF training} gets stuck in local minima. 
We also evaluate the learned radiance field with novel, extrapolated poses.
Our view syntheses results are comparable to those \joycey{trained with} GT poses.
As expected, the quality degrades when the evaluation pose deviates too far from the training poses, especially \joycey{for} the background scene that is unseen in the training videos.
However, we can still see a person standing on the pavement and observe the structure of the scene.
On the other hand, due to the incorrect LoFTR poses, the baseline fails to produce realistic results.

\vspace{-2mm}
\paragraph{Extending VCs to other objects: } 
As a proof-of-concept, we exploit canonical 3D deformable mapping \cite{Novotny2020} as shape priors and adapt our method to cars. 
As shown in Fig. \ref{fig:vc-cars}, we are able to estimate VCs and recover relative poses effectively (pose error: $16^\circ$) even from extreme viewpoints.
We refer the readers to supp. material for more details.

\section{Conclusion}

We introduced a novel concept called virtual correspondences -- a pair of image points whose camera rays intersect in 3D. Unlike classic correspondences, virtual correspondences do not need to describe the same, co-visible 3D points. Thus, VCs are not constrained by visual or semantic similarities, making it possible to match images with little or no overlap. We proposed a method to extract virtual correspondences based on prior knowledge of foreground objects in the image, and integrate with existing 3D frameworks. Our experiments on two challenging human-based datasets show that virtual correspondences are critical towards successful camera pose estimation and downstream multi-view stereo and novel view synthesis in extreme-view scenarios. 

\paragraph{Social impact: }
Our method alleviates the need to capture dense views for camera pose estimation and 3D reconstruction, and has the potential to reduce storage and computational costs. Unfortunately, it could also be exploited by surveillance and may raise privacy concerns as 3D reconstruction from few images becomes more accessible.

\paragraph{Acknowledgment: } We would like to thank Ioan Andrei Bârsan, Siva Manivasagam, and Kelvin Wong for their feedback on the early draft, and Lucy Chai for final proofreading.

{\small
\bibliographystyle{ieee_fullname}
\bibliography{egbib}
}

\end{document}